\documentclass[11pt,a4paper]{article}
\usepackage[nohyperref]{naaclhlt2019}
\usepackage{times}
\usepackage{latexsym}
\usepackage{xcolor}
\usepackage{amsmath, amssymb, amsthm, enumerate, framed, graphicx}
\usepackage[normalem]{ulem}
\usepackage{booktabs}
\usepackage{multirow}

\usepackage{url}

\newcommand{\xv}{\mathbf{x}}
\newcommand{\yv}{\mathbf{y}}
\newcommand{\zv}{\mathbf{z}}
\newcommand{\uv}{\mathbf{u}}
\newcommand{\wm}{\mathbf{W}}
\newcommand{\wst}{\mathbf{W}_{xy}}
\newcommand{\wts}{\mathbf{W}_{yx}}
\newcommand{\fst}{f_{xy}}
\newcommand{\fts}{f_{yx}}
\newcommand{\argmin}{\operatornamewithlimits{argmin}}

\aclfinalcopy 
\def\aclpaperid{1933} 

\title{Density Matching for Bilingual Word Embedding}

\author{Chunting Zhou, Xuezhe Ma, Di Wang, Graham Neubig \\
  Language Technologies Institute \\
  Carnegie Mellon University \\
  {\tt {ctzhou,xuezhem,diwang,gneubig}@cs.cmu.edu} \\}
  
\date{}

\begin{document}
\maketitle
\begin{abstract}
Recent approaches to cross-lingual word embedding have generally been based on linear transformations between the sets of embedding vectors in the two languages. In this paper, we propose an approach that instead expresses the two monolingual embedding spaces as probability densities defined by a Gaussian mixture model, and matches the two densities using a method called normalizing flow. 
The method requires no explicit supervision, and can be learned with only a seed dictionary of words that have identical strings.
We argue that this formulation has several intuitively attractive properties, 
particularly with the respect to improving robustness and generalization to mappings between difficult language pairs or word pairs.
On a benchmark data set of bilingual lexicon induction and cross-lingual word similarity, our approach can achieve competitive or superior performance compared to state-of-the-art published results, with particularly strong results being found on etymologically distant and/or morphologically rich languages.\footnote{Code/scripts can be found at \url{https://github.com/violet-zct/DeMa-BWE}.}
\end{abstract}

\section{Introduction}
Cross-lingual word embeddings represent words in different languages in a single vector space, capturing the syntactic and semantic similarity of words across languages in a way conducive to use in computational models \citep{upadhyay2016cross,ruder2017survey}. These embeddings have been shown to be an effective tool for cross-lingual NLP, e.g. the transfer of models trained on high-resource languages to low-resource ones \citep{klementiev2012inducing,guo2015cross,zoph2016transfer,ahmad2018near,gu2018universal} or unsupervised learning \citep{artetxe2018unsupervised}.

There are two major paradigms in the learning of cross-lingual word embeddings: ``online'' and ``offline''. ``Online'' methods learn the cross-lingual embeddings directly from parallel corpora \citep{hermann2014multilingual}, optionally augmented with monolingual corpora \citep{gouws2015bilbowa}. In contrast, ``offline'' approaches learn a bilingual mapping function or multilingual projections from pre-trained monolingual word embeddings or feature vectors \cite{haghighi2008bilingual,mikolov2013exploiting,faruqui2014multilingual}.
In this work, we focus on this latter offline approach.

\begin{figure}[tb]
  \centering
  \includegraphics[scale=0.33]{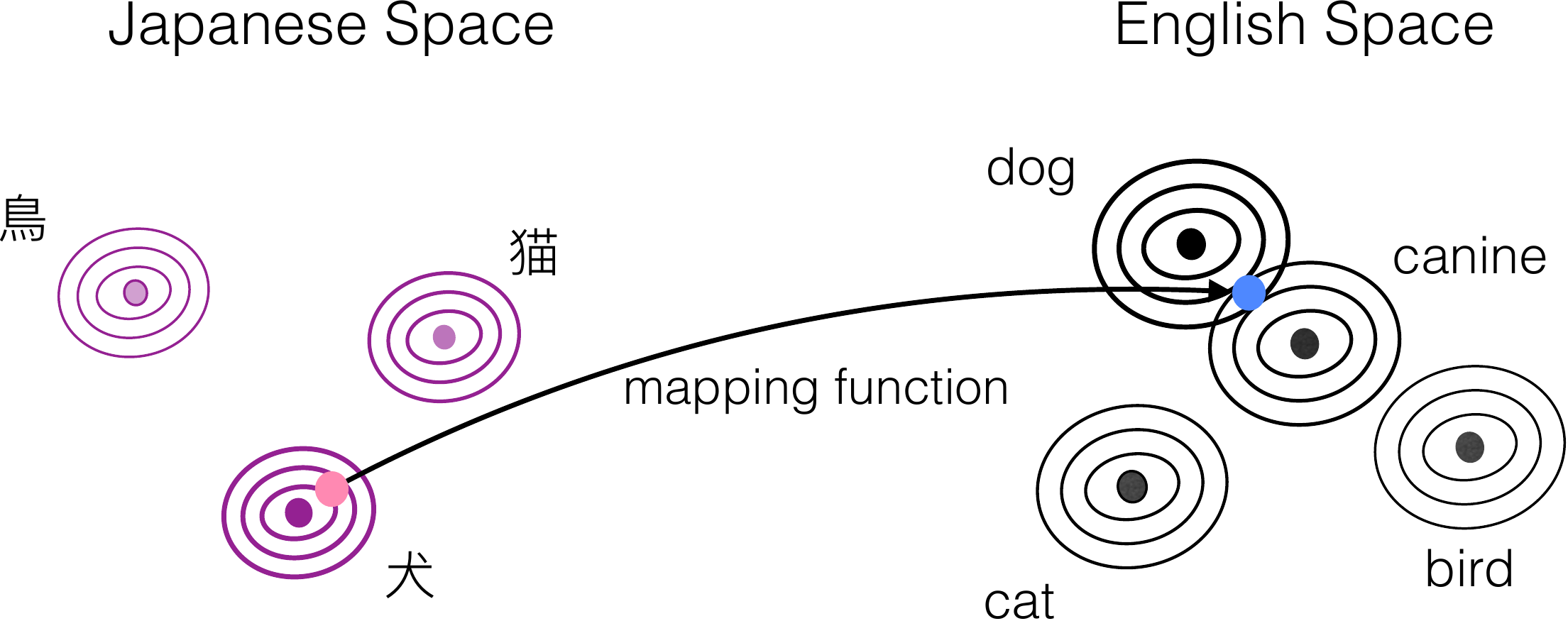}
  \caption{An illustration of our method. Thicker lines represent higher mixture weights, linked to word frequency. The pink point is a continuous training sample in the Japanese embedding space while the blue point is a mapped point in the English space.} \label{fig:diagram}
\end{figure}

The goal of bilingual embedding is to learn a shared embedding space where words possessing similar meanings are projected to nearby points.
Early work focused on supervised methods maximizes the similarity of the embeddings of words that exist in a manually-created dictionary, according to some similarity metric \citep{mikolov2013exploiting,faruqui2014multilingual,jawanpuria2018learning,joulin2018loss}.
In contrast, recently proposed unsupervised methods frame this problem as minimization of some form of distance between the whole set of discrete word vectors in the chosen vocabulary, e.g. Wasserstein distance or Jensen--Shannon divergence \citep{xu2018unsupervised,conneau2017word,zhang2017earth,grave2018unsupervised}.
While these methods have shown impressive results for some language pairs despite the lack of supervision, regarding the embedding space as a set of discrete points has some limitations.
First, expressing embeddings as a single point in the space doesn't take into account the inherent uncertainty involved in learning embeddings, which can cause embedding spaces to differ significantly between training runs \cite{wendlandt2018instability}.
Second, even in a fixed embedding space the points surrounding those of words that actually exist in the pre-trained vocabulary also often are coherent points in the embedding space.

In this work, we propose a method for \textbf{density matching for bilingual word embedding (DeMa-BWE)}.
Instead of treating the embedding space as a collection of discrete points, we express it as a probability density function over the entire continuous space over word vectors. 
We assume each vector in the monolingual embedding space is generated from a Gaussian mixture model with components centered at the pretrained word embeddings (Fig. \ref{fig:diagram}), and our approach then learns a bilingual mapping that most effectively matches the two probability densities of the two monolingual embedding spaces.

To learn in this paradigm, instead of using the pre-trained word embeddings as fixed training samples, at every training step we obtain samples from the Gaussian mixture space.
Thus, our method is exploring the entire embedding space instead of only the specific points assigned for observed words.
To calculate the density of the transformed samples, we use volume-preserving invertible transformations over the target word embeddings, which make it possible to perform density matching in a principled and efficient way \citep{rezende2015variational,papamakarios2017masked,he2018unsupervised}.
We also have three additional ingredients in the model that proved useful in stabilizing training: (1) a back-translation loss to allow the model to learn the mapping jointly in both directions, (2) an identical-word-matching loss that provides weak supervision by encouraging the model to have words with identical spellings be mapped to a similar place in the space, and (3) frequency-matching based Gaussian mixture weights that accounts for the approximate frequencies of aligned words.


Empirical results are strong; our method is able to effectively learn bilingual embeddings that achieve competitive or superior results on the MUSE dataset \citep{conneau2017word} over state-of-the-art published results on bilingual word translation and cross-lingual word similarity tasks.
The results are particularly encouraging on etymologically distant or morphologically rich languages, as our model is able to explore the integration over the embedding space by treating the space as a continuous one.
Moreover, unlike previous unsupervised methods that are usually sensitive to initialization or require sophisticated optimization procedures, our method is robust and requires no special initialization.


\section{Background: Normalizing Flows}
\label{sec:nf}
In this section, we will briefly describe normalizing flows - the backbone of DeMa-BWE.


As mentioned in the introduction and detailed later, our model is based on matching two probability density functions, one representing the source embedding space and one representing the target embedding space.
To learn in this framework, we will use the concept of \emph{normalizing flows} \cite{rezende2015variational}.
We will explain them briefly here, but refer readers to \newcite{rezende2015variational} for details due to space constraints.

Concretely, let $\uv$ denote a high dimensional random vector (e.g. representing a point in the source embedding space) and $\zv$ be a latent variable that corresponds to $\uv$ (e.g. a point in the target embedding space).
Flow-based generative models \citep{kingma2016improved,kingma2018glow} learn invertible transformations $f_{\mathbf{\theta}}(\zv)$ from the distribution over $\zv$ to the distribution over $\uv$.
The generative story of the model is defined as:
\begin{align}
    \zv \sim p_\mathbf{\theta}(\zv), ~~~~ \uv = f_\mathbf{\theta}(\zv)
\end{align}
where $p_\mathbf{\theta}(\zv)$ is the prior distribution.
This prior can be any distribution for which we can tractably compute the density of sample points $\zv$.
A common choice of such distribution is a spherical multivariate Gaussian distribution: $p_\mathbf{\theta}(\zv) = \mathcal{N}(\zv; 0, \mathbf{I})$ \citep{dinh2016density}.
Assuming the transformation function $f_\mathbf{\theta}(\cdot)$ is invertible, using the rule for change of variables, the probability density of $\uv$ can be calculated as:
\begin{align}
    \label{eq:density}
    p_{\mathbf{\theta}}(\uv) &=  p_\mathbf{\theta}(\zv)|\mathrm{det}(J(f^{-1}(\uv)))|
\end{align}
where $\mathrm{det}(J(f^{-1}(\uv)))$ is determinant of the Jacobian matrix of the function inverse. 
This term accounts for the way in which $f$ locally expands or contracts regions of $\zv$, and enforces the invertibility of the function. 
A ``normalizing flow'' is a cascaded sequence of such invertible transformations, which is learned by maximizing the density in Equation (\ref{eq:density}) over observed data points $\uv$. One computational issue with these models lies in calculating the Jacobian matrix, which is expensive in the general case. 
A common method is to choose transformations whose Jacobians' are a triangular matrix, which renders this computation tractable~\citep{dinh2016density,papamakarios2017masked}.


\section{Proposed Method}

In this section, we present notation used in our method, describe the prior we define for the monolingual embedding space, then detail our density matching method.

\subsection{Notation}

Given two sets of independently trained monolingual embeddings,
the problem of bilingual embedding mapping is to learn a mapping function that aligns the two sets in a shared space.
Let $\xv \in \mathbb{R}^d, \yv \in \mathbb{R}^d$ denote vectors in the source and target language embedding space respectively. Let $x_i$ and $y_j$ denote an actual word in the pretrained source and target vocabularies respectively. Words are sorted by their occurrence counts in the monolingual corpus and the index of the word represents its rank. We use $\xv_i$ and $\yv_j$ to denote the pretrained word embeddings for word $x_i$ in the source language and word $y_j$ in the target language respectively.
Given a pair of languages $s$ and $t$, our approach learns two mapping functions: $\fst$ that maps the source embedding to the target space and $\fts$ that gives the mapping in the reverse direction.


\subsection{Density Estimation in Monolingual Space} 
\label{sec:model}
To learn the mapping, we project a vector $\xv$ in the source embedding space into the target space $\yv$. We learn this mapping by maximizing the density of data points in the source space. The density can be computed using the idea of normalizing flow described above. Thus, for the the monolingual embedding spaces, we need to define tractable density functions $p(\xv)$ and $p(\yv)$.

While any number of functions could be conceived to calculate these densities, in the current method we opt to use a Gaussian Mixture Model (GMM) with Gaussian components centered at each pretrained word embedding.
This is motivated by the assumption that embeddings are likely to appear in the neighborhood of other embeddings, where we define the ``neighborhood'' to be characterized as closeness in Euclidean space, and the uncertainty of each neighborhood as being Gaussian. 
Concretely, let $N_x$ and $N_y$ denote the number of pretrained word embeddings that serve as Gaussian component centers during training for the source and target languages, respectively.
Then we can express the density of any point in the source embedding space as:
\begin{equation}
    p(\xv) = \sum_{i \in \{1, \ldots, N_x \}}\pi(x_i)\Tilde{p}(\xv|x_i)
    \label{eq:mog}
\end{equation}
where $\pi(x_i)$ is the frequency of word $x_i$ normalized within the $N_x$ component words, and $\Tilde{p}(\xv|x_i)$ is a Gaussian distribution centered at the embedding of word $x_i$. We simply use a fixed variance $\sigma^2_x$ for all Gaussian components:
\begin{equation}
    \Tilde{p}(\xv|x_i) = \mathcal{N}(\xv|\xv_i, \sigma^2_x\mathbf{I})
    \label{eq:gaussian}
\end{equation}
Similarly, the density of any point in the target embedding space can be written as:
\begin{equation}
    p(\yv) = \sum_{j \in \{1, \ldots, N_y \}}\pi(y_j)\Tilde{p}(\yv|y_j)
    \label{eq:target}
\end{equation}
where $\Tilde{p}(\yv|y_j) = \mathcal{N}(\yv|\yv_j, \sigma^2_y\mathbf{I})$.

\subsection{Density Matching}
With the Gaussian mixture model as the prior distribution in the monolingual space, our goal is to learn a mapping function from one embedding space to the other such that the log probabilistic density is maximized in the source space.

While we are jointly learning the two mapping functions $\fst$ and $\fts$ simultaneously, for conciseness we will illustrate our approach using the source to target mapping $\fst$. 
First, a continuous vector $\xv$ is sampled from the Gaussian mixture model (Eq. \eqref{eq:mog}) by sampling $x_i \sim \pi(x_i)$ then $\xv \sim \Tilde{p}(\xv|x_i)$ \eqref{eq:gaussian}. Next, we apply the mapping function $\fst$ to obtain the transformed vector $\yv$ in the target space.
Concretely, the mapping functions we employ in this work are two linear transformations: $\fst(\cdot) = \wst\cdot$ and $\fts = \wts\cdot$.
Connecting to the transformation function in Sec. \ref{sec:nf}, we see that $\xv = f(\yv) = \wst^{-1}\yv$, $\yv = f^{-1}(\xv) = \wst\xv$, and $J(f^{-1}(\xv))=\wst$.
We can then express the log density of a sample $\xv$ as:
\begin{equation}
    \log p(\xv; \wst) = \log p(\yv) + \log \big | \mathrm{det}(\wst) \big |
    \label{eq:target-density}
\end{equation}
where the Jacobian regularization term accounts for the volume expansion or contraction resulting from the projection matrix $\wst$. 

We maximize the likelihood function which is equivalent to minimizing expectation of the KL-divergence between the prior distribution and the model distribution of $\xv$. This provides a natural objective for optimization:
\begin{equation}
    \text{minimize:~~} \mathrm{KL}(p(\xv) || p(\xv; \wst))
\end{equation}
By replacing $\wst\xv$ with $\yv$, this is equivalent to maximizing the log density of transformed source samples in the target space (see Eq. \eqref{eq:target-density}):
\begin{equation}
    \mathcal{L}_{xy} =  \mathrm{E}_{\xv \sim p(\xv)}[ \log p(\yv) + \log \big | \mathrm{det}(\wst) \big | ]
    \label{eq:DensityObj}
\end{equation}
The objective $\mathcal{L}_{xy}$ contains two parts: the log density function $\log p(\yv)$ and a regularization term $\log\mathrm{det}(\wst)$.
Likewise, for the target to source mapping $\wts$, we have the density matching objective $\mathcal{L}_{yx}$.

\paragraph{Conditional Density Matching}
The above marginal density matching method does not take into account the dependency between the embeddings in the two monolingual spaces.
To address this issue, we extend the density matching method to the conditional density function:
\begin{equation}
    \log p(\xv|x_i; \wst) = \log p(\yv |x_i) + \log \big | \mathrm{det}(\wst) \big | \nonumber
\end{equation}
The conditional density $p(\yv |x_i)$ is the prior distribution in this simple normalizing flow, and for this we use a Gaussian mixture model in the target monolingual space:
\begin{align}
\small
 \nonumber   p(\yv | x_i) &= \sum_{j \in \{1, \ldots, N_y \}} p(\yv, y_j|x_i) \\
                 &= \sum_{j \in \{1, \ldots, N_y \}} \Tilde{p}(\yv|y_j)\pi(y_j|x_i)
    \label{eq:target-mixture}
\end{align}
\normalsize
Similarly, 
where $\Tilde{p}(\yv|y_j)$ is the Gaussian density function in the mixture model defined in Equation \eqref{eq:target}. 
$\pi(y_j|x_i)$ allows us to incorporate a-priori assumptions about whether two words are likely to match.
In fact, the density matching method in Eq.~\eqref{eq:target-density} can be regarded as a special case of the conditional density matching method by adopting a naive prior $\pi(y_j|x_i) := \pi(y_j)$.
However previous work \cite{zhang2017earth} has noted that word frequency information is a strong signal -- words with similar frequency are likely to be matched -- and thus we use $\pi(y_j|x_i)$ to incorporate this prior knowledge.

In this work, we assume that the frequencies of aligned bilingual words in the individual monolingual corpus should be correlated, and to match words that are ranked similarly, we model the Gaussian mixture weights as the negative absolute difference between log-scale word ranks and normalize over all the target Gaussian component words by a $\mathrm{softmax}$ function with temperature $\tau$:
\begin{equation}
    \pi(y_j|x_i) = \frac{\exp(-|\log (j) - \log (i)|)/\tau}{\sum_{k=1}^{N_y}\exp(-|\log (k) - \log (i)|/\tau)} \nonumber
\end{equation}
Thus, if a word $x_i$ has similarly frequency rank as word $y_j$, the sample $\xv$ from the Gaussian distribution centered at $\xv_i$ will be assigned higher weight for the component $\Tilde{p}(\yv|y_j) = \Tilde{p}(\wst\xv|y_j)$. Although this assumption will not hold always (e.g. for languages that have different levels of morphological complexity), intuitively we expect that using this signal will help more overall than it will hurt, and empirically we find that this weighting is not sensitive to language variation and works well in practice.

The updated objective is
\begin{align}
    \mathcal{L}_{xy} & =  \mathrm{E}_{x_i\sim\pi(x_i)}[\mathrm{KL}(\Tilde{p}(\xv|x_i) || p(\xv|x_i; \wst))] \nonumber \\
    & = \mathrm{E}_{x_i\sim\pi(x_i), \xv \sim \Tilde{p}(\xv|x_i)}\big [ \log p(\yv |x_i) \\ &+ \log \big | \mathrm{det}(\wst) \big | \big ] \nonumber
\end{align}
In the conditional density above, both the frequency-matching weight and the Gaussian density function play an important role in matching the density of a source-space sample with the target embedding space. The former matches bilingual words with their frequency ranks while the latter matches words with their vector distances.

\subsection{Weak Orthogonality Constraint}
A common choice of bilingual mapping function is a linear transformation with an orthogonality constraint.
Various motivations have been proposed for the orthogonality constraint such as length normalization of embeddings \citep{xing2015normalized}, and reversible mapping \citep{smith2017offline}.
In this work, we add a weak orthogonality constraint to the bilingual mappings via a back-translation loss as follows:
\begin{align}
  \nonumber \mathcal{L}_{bt} &= \mathrm{E}_{x_i\sim\pi(x_i), \xv \sim \Tilde{p}(\xv|x_i)}\big [ g(\wts\wst\xv, \xv)\big ] \\ &+ \mathrm{E}_{y_j\sim\pi(y_j), \yv \sim \Tilde{p}(\yv|x_j)}\big [g(\wst\wts\yv, \yv) \big ] \nonumber
\end{align}
where $g(\cdot, \cdot) = 1 - \mathrm{cosine}(\cdot, \cdot)$ is the cosine loss. Jointly learning the two mapping matrices by minimizing this cyclic loss encourages $\wst$ and $\wts$ to be orthogonal to each other.
\subsection{Weak Supervision with Identical Words}
To reduce the search space of the mapped bilingual embeddings, we add an additional weakly supervised loss over words that have identical strings in both the source and target languages denoted $\mathcal{W}_{id}$.
\begin{equation}
        \mathcal{L}_{sup} = \sum_{v\in{\mathcal{W}_{id}}}{g(\mathbf{v}_x\wst^T, \mathbf{v}_y) + g(\mathbf{v}_y\wts^T, \mathbf{v}_x)} \nonumber
\end{equation}
where $\mathbf{v}_x$ and $\mathbf{v}_y$ are the pretrained word embedding of word $v$ in the source and target side respectively, and $g(\cdot, \cdot)$ is the cosine loss described above.
Although using identical strings for supervision is very noisy, especially for languages with little overlap in vocabularies, we find that they provide a enough guidance to training to prevent the model from being trapped in poor local optima.

Putting everything together, the overall objective function of DeMa-BWE includes three parts: the density matching loss, the weak orthogonality loss and the weak supervised loss:
\begin{align}
   \mathcal{L} = \mathcal{L}_{xy} + \mathcal{L}_{yx} + \lambda \cdot \mathcal{L}_{bt} + \alpha \cdot \mathcal{L}_{sup} 
\end{align}
where $\lambda$ and $\alpha$ are coefficients that tradeoff between different losses.
\section{Retrieval and Refinement}
\subsection{Retrieval Method}
One standard use case for bilingual embeddings is in bilingual lexicon induction, where the embeddings are used to select the most likely translation in the other language given these embeddings.
In this case, it is necessary to have a retrieval metric that selects word or words likely to be translations given these embeddings.
When performing this retrieval, it has been noted that
high-dimensional embedding spaces tend to suffer from the ``hubness" problem \cite{Radovanovic2010Hubness} where some vectors (known as \textit{hubs}) are nearest neighbors of many other points, which is detrimental to reliably retrieving translations in the bilingual space. To mitigate the hubness problem, we adopt the \textit{Cross-Domain Similarity Local Scaling} (CSLS) metric proposed in \citep{conneau2017word} that penalizes the similarity score of the hubs. Specifically, given two mapped embeddings $\wst\xv$ denoted $\xv'$ and $\yv$, CSLS first computes the average cosine similarity of $\xv'$ and $\yv$ for their $k$ nearest neighbors denoted $r_T(\xv')$ and $r_S(\yv)$ in the other language respectively, then the corrected similarity measure $\mathrm{CSLS}(\cdot, \cdot)$ is defined as:
\begin{align}
    \mathrm{CSLS}(\xv', \yv) = 2\mathrm{cos}(\xv', \yv) - r_T(\xv') - r_S(\yv) \nonumber
\label{eq:csls}
\end{align}
where $\mathrm{cos}(\cdot, \cdot)$ is the cosine similarity. Following \citep{conneau2017word}, $k$ is set to be 10.

CSLS consistently outperform cosine similarity on nearest neighbor retrieval, however it does not consider the relative frequency of bilingual words which we hypothesize can be useful in disambiguation.
As we discussed in Sec. \ref{sec:model}, our density matching objective considers both the relative frequencies and vector similarities.
The conditional density $p(\yv|x_i)$ (Eq. \eqref{eq:target-mixture}) in our density matching objective (Eq. \eqref{eq:DensityObj}) is a marginalized distribution over all target component words $y_j$ where the density of each component $p(\yv, y_j|x_i)$ can be directly used as a similarity score for a pair of words ($y_j$, $x_i$) to replace the cosine similarity $\mathrm{cos}(\xv', \yv)$ in CSLS. Let CSLS-D denote this modified CSLS metric, which we compare to standard CSLS in experiments. 
We find that using CSLS-D for nearest neighbor retrieval outperforms the CSLS metric in most cases on the bilingual dictionary induction task.

\subsection{Iterative Procrustes Refinement}
\label{sec:refine}


Iterative refinement, which learns the new mapping matrix by constructing a bilingual lexicon iteratively, has been shown as an effective method for improving the performance of unsupervised lexicon induction models \cite{conneau2017word}.
Given a learned bilingual embedding mapping $\wm$, the refinement starts by inferring an initial bilingual dictionary using the retrieval method above on the most frequent words. Let $\mathbf{X}$ and $\mathbf{Y}$ denote the ordered embedding matrices for the inferred dictionary words for source and target languages respectively. Then a new mapping matrix $\wm^*$ is induced by solving the Procrustes problem:
\begin{equation}
 \nonumber   \begin{array}{c}
    \wm^* = \argmin\limits_{\wm\in O_d(\mathbb{R})} ||\wm\mathbf{X} - \mathbf{Y}||_F= \mathbf{U}\mathbf{V}^T  \\
     s.t. ~ \mathbf{U}\Sigma\mathbf{V}^T = \mathrm{SVD}(\mathbf{Y}\mathbf{X}^T)
    \end{array}
\end{equation}
The step above can be applied iteratively by inducing a new dictionary with the new mapping $\wm$.

DeMa-BWE is able to achieve very competitive performance without further refinement, but for comparison we also report results with the refinement procedure, which brings small improvements in accuracy for most language pairs.
Note that for bilingual dictionary induction during refinement, we use CSLS as the retrieval metric across all experiments for fair comparison to the refinement step in previous work.

\section{Experiments}
\subsection{Dataset and Task} 
We evaluate our approach extensively on the bilingual lexicon induction (BLI) task, which measures the word translation accuracy in comparison to a gold standard. We report results on the widely used MUSE dataset \citep{conneau2017word}, which consists of FastText monolingual embeddings pre-trained on Wikipedia \citep{bojanowski2017enriching}, and dictionaries for many language pairs divided into train and test sets. We follow the evaluation setups of \citep{conneau2017word}. We evaluate DeMa-BWE by inducing lexicons between English and different languages including related languages, e.g. Spanish; etymologically distant languages, e.g. Japanese; and morphologically rich languages, e.g. Finnish.

\begin{table*}[t]
\resizebox{\textwidth}{!}{%
\begin{tabular}{@{}lrrrrrrrrrrrr@{}}
\toprule
& en-es & es-en & en-de & de-en & en-fr & fr-en & en-ru & ru-en & en-zh & zh-en & en-ja & ja-en \\ \midrule
\multicolumn{13}{c}{Supervised} \\ \midrule
Procrustes (R) & 81.4 & 82.9 & 73.5 & 72.4 & 81.1 & 82.4 & 51.7 & 63.7 & 42.7 & 36.7 & 14.2 & 7.44 \\
MSF-ISF & 79.9 & 82.1 & 73.0 & 72.0 & 80.4 & 81.4 & 50.0 & 65.3 & 28.0 & 40.7 & - & - \\
MSF & 80.5 & 83.8 & 73.5 & 73.5 & 80.5 & 83.1 & 50.5 & 67.3 & 32.3 & 43.4 & - & - \\
CSLS-Sp & \textit{84.1} & \textit{86.3} & \textit{79.1} & 76.3 & \textit{83.3} & \textit{84.1} & \textit{57.9} & 67.2 & 45.9 & \textit{46.4} & - & - \\
GeoMM & 81.4 & 85.5 & 74.7 & \textit{76.7} & 82.1 & \textit{84.1} & 51.3 & \textit{67.6} & \textit{49.1} & 45.3 & - & - \\ \midrule\midrule
\multicolumn{13}{c}{Unsupervised} \\ \midrule
MUSE (U+R) & 81.7 & 83.3 & 74.0 & 72.2 & 82.3 & 81.1 & 44.0 & 59.1 & 32.5 & 31.4 & 0.0 & 4.2 \\
SL-unsup & 82.3 & 84.7 & 75.1 & 74.3 & 82.3 & 83.6 & 49.2 & 65.6 & 0.0 & 0.0 & 2.9 & 0.2 \\
SL-unsup-ID & 82.3 & 84.6 & 75.1 & 74.1 & 82.2 & \textbf{83.7} & 48.8 & \textbf{65.7} & 37.4 & 34.2 & 48.5 & 33.7 \\
$\text{Sinkhorn}^*$ & 79.5 & 77.8 & 69.3 & 67.0 & 77.9 & 75.5 & - & - & - & - & - & - \\
Non-Adv & 81.1 & 82.1 & 73.7 & 72.7 & 81.5 & 81.3 & 44.4 & 55.6 & 0.0 & 0.0 & 0.0 & 0.0 \\
Non-Adv (R) & 82.1 & 84.1 & 74.7 & 73.0 & 82.3 & 82.9 & 47.5 & 61.8 & 0.0 & 0.0 & 0.0 & 0.0 \\
WS-Procrustes (R) & 82.8 & 84.1 & 75.4 & 73.3 & 82.6 & 82.9 & 43.7 & 59.1 & - & - & - & - \\ \midrule
\multicolumn{13}{c}{DeMa-BME} \\ \midrule
CSLS (w/o R) & 82.0 & \textbf{85.4} & 75.3 & 74.9 & 82.6 & 82.4 & 46.9 & 62.4 & 39.6 & \textbf{40.0} & 46.7 & 32.9 \\
CSLS-D (w/o R) & 82.3 & 85.1 & 76.3 & \textbf{75.1} & \textbf{83.2} & 82.5 & 48.0 & 61.7 & 40.5 & 37.7 & 45.3 & 32.4 \\
CSLS (w/ R) & \textbf{82.8} & 84.5 & 75.6 & 74.1 & 82.5 & 83.3 & 47.3 & 63.5 & 41.9 & 37.7 & 50.7 & 35.2 \\
CSLS-D (w/ R) & \textbf{82.8} & 84.9 & \textbf{77.2} & 74.4 & 83.1 & 83.5 & \textbf{49.2} & 63.6 & \textbf{42.5} & 37.9 & \textit{\textbf{52.0}} & \textit{\textbf{35.6}} \\ \bottomrule
\end{tabular}
}
\caption{Precision@1 for the MUSE BLI task compared with previous work.
All the baseline results employ CSLS as the retrieval metric except for $\text{Sinkhorn}^*$ which uses cosine similarity.
\textbf{R} represents refinement. Bold and italic indicate the best unsupervised and overall numbers respectively. ('en' is English, 'es' is Spanish, 'de' is German, 'fr' is French, 'ru' is Russian, 'zh' is traditional Chinese, 'ja' is Japanese.)}
\label{tab:main}
\vspace{-3mm}
\end{table*}

\subsection{Implementation Details}
\paragraph{Embedding Normalization}
Following \citep{artetxe2018robust}, we pre-process the monolingual embeddings by first applying length normalization, then mean center each dimension, and then length normalize again to ensure that the final embeddings have a unit length. We observe that this normalization method helps stabilize training and accelerate convergence.

\paragraph{Other Experimental Details}
We held out 1000 translation pairs randomly sampled from the training set in the MUSE dataset as our validation data. We also tried the unsupervised validation criterion proposed in \cite{conneau2017word} as the model selection method that computes the average cosine similarity over the model induced dictionary pairs and found that this unsupervised criterion can select models that achieve similar performance as the supervised validation criterion. All hyperparameters are tuned on the validation set and include the following: For the number of base words used as Gaussian components in the GMM, we typically choose the most frequent 20,000 words for all language pairs 
but en-ja for which we use 10,000 which achieves better performance.
We use a batch size of 2048 for all languages but en-ja for which we use 1024. 
We use Adam \cite{kingma2014adam} for optimization with default hyperparameters.

We empirically set the Gaussian variance to be 0.01 for both the source and target languages in en-es, en-de, en-fr, en-ru; in the experiments for morphologically rich languages (Sec. \ref{sec:morph}), we set the variance to be 0.015 for all these languages except for et whose variance is set to be 0.02 while keeping the variance of English to be 0.01. 
In the experiments for etymologically distant language pairs en-ja and en-zh, we set different variances for the source and target languages in different mapping directions. For details of the variance setting please check the scripts in our code base.
We empirically find that for a language pair the variance of the language with relatively more complex morphological properties needed to be set larger than the other language, indicating that the model needs to explore more in the embedding space for the morphologically richer language.

We initialize mapping matrices $\wst$ and $\wts$ with a random orthogonal matrix. For the weak orthogonality constraint loss $\mathcal{L}_{bt}$, we set the weight $\lambda$ to be 0.5 throughout all language pairs.
For the weak supervision loss $\mathcal{L}_{sup}$, we set the weight $\alpha$ to be 10 for all languages except for en-zh where we find 5 performs better.
We set the temperature $\tau$ used in the $\mathrm{softmax}$ function for Gaussian mixture weights to be 2 across all languages.
\begin{table*}[t]
\resizebox{\textwidth}{!}{%
\begin{tabular}{lrrrrrrrrrrrr}
\toprule
                             & en-et & et-en & en-fi & fi-en & en-el & el-en & en-hu &          hu-en & en-pl & pl-en & en-tr & tr-en \\ \midrule
MUSE (U+R)                   &  1.7 &  0.1 &  0.1 & 59.8 & 39.1 & 59.0 & 50.2 &           0.1 & 53.9 &  0.0 & 45.4 &  0.0 \\
5k+Procrustes (R)            & 31.9 & 45.6 & 47.3 & 59.5 & 44.6 & 58.5 & 53.3 & 64.8 & 58.2 & 66.9 & 46.3 & 59.2 \\
id+Procrustes (R)            & 29.7 & 40.6 & 45.0 & 59.1 & 40.7 & 55.1 & 52.6 &          63.7 & 57.3 & 66.7 & 45.4 & 61.4 \\
$\text{id+Procrustes (R)}^*$ & 31.5 &     - & 28.0 &     - & 42.9 &     - & 46.6 &              - & 52.6 &     - & 39.2 &     - \\ \midrule
CSLS (w/o R)                 & 32.9 & 45.3 & 47.5 & 58.7 & 43.6 & 57.8 & 55.3 &          64.5 & 59.9 & 69.1 & 50.3 & 60.8 \\
CSLS-D (w/o R)               & 35.9 & 45.2 & 49.4 & 58.5 & 45.0 & 58.1 & 57.6 &          64.7 & 60.9 & 68.3 & 52.8 & \textbf{61.6} \\
CSLS (w/ R)                  & 34.4 & 47.8 & 50.2 & 60.5 & 44.5 & 61.1 & 56.4 &          64.8 & 59.7 & 69.0 & 50.3 & 60.8 \\
CSLS-D (w/ R)                & \textbf{37.0} & \textbf{47.9} & \textbf{52.4} & \textbf{60.8} & \textbf{46.3} & \textbf{61.6} & \textbf{59.2} & \textbf{64.9} & \textbf{61.5} & \textbf{69.1} & \textbf{53.4} & 61.2 \\ \bottomrule
\end{tabular}%
}
\caption{BLI Precision (@1) for morphologically complex languages. \textbf{$\text{id+Procrustes (R)}^*$} is the result reported in \citep{sogard2018limitations}. \textbf{5k+Procrustes (R)} uses the training dictionary with 5k unique query words. ('et' is Estonian, 'fi' is Finnish, 'el' is Greek, 'hu' is Hungarian, 'pl' is Persian, 'tr' is Turkish.)}
\label{tab:morph}
 \vspace{-2mm}
\end{table*}

\subsection{Main Results on BLI}
In Tab. \ref{tab:main}, we compare the performance of DeMa-BME extensively with the best performing unsupervised and supervised methods on the commonly benchmarked language pairs.

Our unsupervised baselines are: (1) \textbf{MUSE (U+R)} \citep{conneau2017word}, a GAN-based unsupervised method with refinement. (2) A strong and robust unsupervised self-learning method \textbf{SL-unsup} from \citep{artetxe2018robust}. We also run their published code with identical words as the initial dictionary for fair comparison with our approach, denoted \textbf{SL-unsup-ID}. 
(3) \textbf{Sinkhorn} \citep{xu2018unsupervised} that minimizes the Sinkhorn distance between the source and target word vectors. (4) An iterative matching method from \citep{hoshen2018non}): \textbf{Non-Adv} and \textbf{Non-Adv (R)} with refinement. (5) \textbf{WS-Procrustes (R)} using refinement by \citep{grave2018unsupervised}. Our supervised methods include: (1) The iterative Procrustes method \textbf{Procrustes (R)} \citep{smith2017offline}.
(2) A multi-step framework \textbf{MSF-ISF} \citep{artetxe2018generalizing} and its variant \textbf{MSF} which uses CSLS for retrieval, whose results are from \citep{jawanpuria2018learning}. (3) \textbf{CSLS-Sp} by \citep{joulin2018loss} that optimizes the CSLS score, and (4) a geometric approach \textbf{GeoMM}
by \citep{jawanpuria2018learning}. For fair comparisons, all supervised results are trained with the training dictionaries in the MUSE dataset.
All baseline methods employ CSLS for retrieval except for the Sinkhorn method.

For DeMa-BME, we present results with and without refinement, and with CSLS and CSLS-D as retrieval methods. From Tab. \ref{tab:main}, we can see the overall performance of DeMa-BME is remarkable comparing with other unsupervised methods and is also competitive with strong supervised methods.
The results without the iterative refinement \textbf{CSLS (w/o R)} are strong on almost all language pairs with particularly strong performance being observed on es-en, en-de and en-fr on which DeMa-BME outperforms or is on par with the best performing methods.
Applying refinement to DeMa-BME brings slightly better performance on most language pairs but degrades the performance on some language pairs such as es-en, zh-en for which the DeMa-BME already obtains very good results.

Also, DeMa-BME demonstrates notably better performance on distant language pairs (en-ru, en-ja and en-zh) over other unsupervised methods, which often achieve good performance on etymologically close languages but fail to converge on the distant language pairs.
However, when the dictionary is initialized with identical strings for SL-unsup, we obtain decent results on these languages.
The strong performance of supervised methods on Russian and Chinese demonstrates that on some language pairs supervised seed lexicons are still necessary.

Finally, when our density-based metric CSLS-D is employed for retrieval, it could achieve further gains in accuracy for most language pairs compared to its counterpart.

\subsection{Morphologically Rich Language Results}
\label{sec:morph}
\citet{sogard2018limitations} found that the commonly benchmarked languages are morphologically poor isolating or exclusively concatenating languages. They select several languages with different morphological traits and complexity then studied the impacts of language similarities on the bilingual lexicon induction.

They show that a simple trick, harvesting the identical word strings in both languages as an initial dictionary and running the iterative Procrustes analysis described in Sec. \ref{sec:refine}, enables more robust and competitive bilingual induction results over the GAN-based unsupervised method with refinement. We denote this method `id+Procrustes (R)'.

Tab. \ref{tab:morph} shows results on the morphologically complex languages used by \citet{sogard2018limitations}. For each language pair we run experiments in both directions.
The baseline methods in \citet{sogard2018limitations} include id+Procrustes (R) and the MUSE (U+R). We run id+Procrustes (R) ourselves and obtain different results from them: except for en-et and en-el, we obtain significantly better results on other language pairs.
In addition, we add another strong supervised baseline (5k+Procrustes (R)) with the training dictionary in the MUSE dataset and iterative Procrustes refinement. From Tab. \ref{tab:morph}, we observe that even without refinement, DeMa-BME (CSLS (w/o R)) outperforms both the unsupervised and supervised baselines on even these difficult morphologically rich languages.

\begin{table}[!h]
\resizebox{\linewidth}{!}{%
\begin{tabular}{@{}lllll@{}}
\toprule
Supervised       & de-en          & es-en          & fa-en          & it-en          \\ \midrule
\citet{xing2015normalized}             & 72          & 71       & 69         & 72          \\
\citet{shigeto2015ridge}          & 72          & \textit{72} & 69          & 72          \\
\citet{artetxe2016learning}        & \textit{73} & \textit{72} & \textit{70} & \textit{73} \\
\citet{artetxe2017learning}       & 70         & 70          & 67          & 71          \\ \midrule 
Unsupervised     &         &          &           &                                  \\ \midrule
\citet{conneau2017word}       & 71          & 71          & 68          & 71         \\
\citet{xu2018unsupervised}         & 71          & 71          & 67          & 71          \\ 
DeMa-BME (w/o R) & \textbf{72.2} & \textbf{72.2} & \textbf{68.6} & \textbf{72.2} \\ 
\bottomrule
\end{tabular}%
}
\caption{Pearson rank correlation ($\times 100$) on cross-lingual word similarity task. Bold indicates the best unsupervised numbers.}
\label{tab:sim}
 \vspace{-2mm}
\end{table}

\subsection{Cross-lingual Word Similarity}
We evaluate DeMa-BWE on the cross-lingual word similarity task from SemEval 2017 \citep{camacho2016nasari} and compare with some strong baselines in \citet{xu2018unsupervised}. In Tab. 5, following the convention in benchmark evaluation for this task, we report the Pearson correlation scores ($\times 100$). DeMa-BME achieves the best performance among all the listed unsupervised methods. Compared with the supervised methods, DeMa-BME is also very competitive. 

\begin{table}[h]
\resizebox{\linewidth}{!}{%
\begin{tabular}{@{}lllll@{}}
\toprule
                        & en-fr & fr-en & en-ja & ja-en \\ \midrule
w/o $L_{xy}$ \& $L_{yx}$& 45.4  & 73.3 & 6.3  & 27.3 \\
w/o $L_{bt}$               & 82.3 & 82.3 & 46.2 & 31.1 \\
w/o $L_{sup}$              & 0.1  & 0.1  & 0.0  & 0.0  \\
$\pi(y_j|x_i) := \pi(y_j)$ & 82.1 & 82.1 & 46.0 & 32.5 \\
Full Model              & 82.6 & 82.4 & 46.7 & 32.9 \\ \bottomrule
\end{tabular}%
}
\caption{Ablation study on different components of DeMa-BME.}
\label{tab:ablation}
\vspace{-2mm}
\end{table}

\subsection{Ablation Study}
Finally, we perform ablation studies in Tab. \ref{tab:ablation} to examine the effects of different components of DeMa-BWE. In comparison to the full model, we remove the density matching loss $L_{xy}$ \& $L_{yx}$, the weakly supervised loss $L_{sup}$, the back-translation loss $L_{bt}$ respectively. First, we observe that without the identical strings as the supervised loss, DeMa-BWE fails to converge as the density matching is difficult given a high-dimensional embedding space to search. Second, when we remove the proposed density matching loss, the model is able to produce reasonable accuracy for fr-en and ja-en, but has undesirable results on en-fr and en-ja, which verifies the necessity of the unsupervised density matching. Third, the back-translation loss is not a crucial component in DeMa-BME; removing it only degrades the model's performance by a small margin. This indicates that orthogonality is not must-have constraint given the model has enough capacity to learn a good transformation.

In addition, we also compare the frequency-matching based Gaussian mixture weights in \eqref{eq:target-mixture} with the naive target frequency based weights. As shown in the fourth row of Tab. \ref{tab:ablation}, the performance of DeMa-BWE with the naive weights is nominally worse than the model using the frequency-matching based mixture weights.


\section{Conclusion}
In this work, we propose a density matching based unsupervised method for learning bilingual word embedding mappings. DeMa-BWE performs well in the task of bilingual lexicon induction. In the future work, we will integrate Gaussian embeddings \citep{vilnis2014word} with our approach.

\section*{Acknowledgments}
This work is sponsored by Defense Advanced Research Projects Agency Information Innovation Office (I2O), Program: Low Resource Languages for Emergent Incidents (LORELEI), issued by DARPA/I2O under Contract No. HR0011-15-C-0114. 
The authors thank Amazon for their gift of AWS cloud credits.
The authors would also like to thank Ruochen Xu, Barun Patra, Joel Ruben Antony Moni for their helpful discussions during drafting this paper.

\bibliography{naaclhlt2019}
\bibliographystyle{acl_natbib}




\end{document}